\documentclass[]{article}

\usepackage[preprint, nonatbib]{neurips_2022}
\usepackage{amssymb}
\usepackage{amsmath}

\usepackage[utf8]{inputenc} 
\usepackage[T1]{fontenc}    
\usepackage{hyperref}       
\usepackage{url}            
\usepackage{booktabs}       
\usepackage{amsfonts}       
\usepackage{nicefrac}       
\usepackage{microtype}      
\usepackage{xcolor}         

\usepackage{amsmath}
\usepackage{graphicx}
\usepackage[numbers]{natbib}
\usepackage{float}
\usepackage{enumitem}
\usepackage{subcaption}
\usepackage{graphicx}

\title{Items or Relations - what do Artificial Neural Networks learn?}

\author{
  Renate Krause$^{1,2,*}$, Stefan Reimann$^{1,*}$  \\
  \\
  1) Institute of Neuroinformatics, University of Zurich and ETH Zurich, Zurich, Switzerland\\
  2) Neuroscience Center Zurich, University of Zurich and ETH Zurich, Zurich, Switzerland\\
  *) equal contribution\\
  Corresponding authors:  rekrau | reinmannst@ini.uzh.ch\\
}

\begin{document}

\maketitle

\begin{abstract}
What has an Artificial Neural Network (ANN) learned after being successfully trained to solve a task - the set of training items or the relations between them? This question is difficult to answer for modern applied ANNs because of their enormous size and complexity.
Therefore, here we consider a low-dimensional network and a simple task, i.e., the network has to reproduce a set of training items identically. 
We construct the family of solutions analytically and use standard learning algorithms to obtain numerical solutions. These numerical solutions differ depending on the optimization algorithm and the weight initialization and are shown to be particular members of the family of analytical solutions. 
In this simple setting, we observe that the general structure of the network weights represents the training set's symmetry group, i.e., the relations between training items. As a consequence, linear networks generalize, i.e., reproduce items that were not part of the training set but are consistent with the symmetry of the training set. In contrast, non-linear networks tend to learn individual training items and show associative memory. At the same time, their ability to generalize is limited. A higher degree of generalization is obtained for networks whose activation function contains a linear regime, such as $\tanh$. 
Our results suggest ANN's ability to generalize - instead of learning items - could be improved by generating a sufficiently big set of elementary operations to represent relations and strongly depends on the applied non-linearity.
~\\

\end{abstract}

\section{Introduction}
Artificial Neural Networks (ANNs) are a valuable tool for neuroscience (\cite{yang2020artificial,bassett2018nature, barak2017recurrent, sporns2014contributions}). They are often successfully trained on neuroscience-inspired tasks and reverse-engineered to generate new hypotheses about the underlying neural computations. Significant effort has been put into analyzing the resulting networks for individual tasks (\cite{sussillo2013opening, mante2013context}). However, it remains challenging to understand what exactly an ANN has learned after being successfully trained to solve a task. 

It is of particular interest to understand whether ANNs explicitly learn training items or rather a set of relations or {concepts} between them. For example, Hopfield's auto-associative networks learn several items as stable point attractors, allowing pattern retrieval and completion. At the same time, they do not generalize, e.g., make predictions or classify new points (\cite{krotov2023new,hopfield1999brain}). On the other hand, learning \textit{concepts} is essential for generalization. 

However, in the currently used ANNs, understanding What-is-Learned is often prevented by the network's complexity, e.g., its architectures and particularly the sheer amount of parameters, ranging from $10^4$ up to $10^8$ and even beyond. Consequently, the main focus is often on developing computationally efficient algorithms to obtain sufficiently good solutions even in very high dimensional parameter space and on their performance in specific tasks.
To achieve good performance, explicit learning of the training items is generally undesirable. It is often referred to as \textit{overfitting} whereby the networks can solve all training examples well but fail to generalize to any new examples presented in a test set. 
Various methods have been proposed to avoid overfitting and are regularly applied to train ANNs more robustly (e.g., dropout \cite{srivastava2014dropout}, L2-regularization \cite{drucker1992improving,krogh1991simple}). These methods generally improve the network's generalization ability but provide little insight into what causes a network to learn training items instead of their relations.

Instead, in this note, we aim to improve our understanding of ANNs by studying the simplest possible architecture: a two-layered network, one layer for input and the other one for output. Moreover, we chose a very simple task: Identically reproduce a set of training items. This auto-association task allows us to gain insight into how the ANNs represent the provided training items in a compressed manner, an essential building block of several more sophisticated ANN tasks.
For concreteness, we construct the training and test set to have a particular set of relations: their symmetry group, i.e., the set of all operations that leave this set invariant. Within this concept, we can clearly distinguish between {\it items}, elements of the training set, and their {\it relations}, elements of the symmetry group of the training set. 

The example is chosen to compare analytical results and numerical estimates resulting from learning algorithms. The more general mathematical setting, including group theory and dynamical systems theory, can be found in \cite{reimann1998design}. In this note, we show that the network connectivities represent the relation between training items over a large set of different optimizers and network hyperparameters. However, the network dynamics can only learn the relations between the items if the network is linear. Non-linear networks learn items. We show in an illustrative example how the transition from linear to non-linear leads the networks from learning relations to learning items. Overall, the presented examples only exemplify the theoretical results which hold for arbitrary network size.

\section{Analytical prelude}\label{sec:results}

In contrast to ANNs used in most applications, the networks we consider in this note are small. Our networks consist of an input layer and an output layer, each of size $3$ without a hidden layer. The output $y$ is a non-linear function of the weighted sum inputs $x$, i.e., $y = \varphi\left( W \: x\right)$, where $W$ is the coupling matrix of the network and $\varphi$ the activation function. The latter is assumed to be strictly monotonously increasing (e.g., linear, sigmoid, or $\tanh$). For convenience, we write $\varphi_W(x):= \varphi(W \: x)$. The theory outlined holds analogously to networks with any particular architecture, such as hidden layers.

We study these networks on an {\bf auto-associative task} in which the networks are asked to reproduce a given set of inputs identically:
\begin{equation}
    x = \varphi_W(x) \qquad  \mbox{for all }x \in X
\end{equation}
This network is called an auto-associator on $X$. If the activation function is the identity $\varphi(x)=x$, then $W={\mathbf I}$ is a trivial solution for any training set. Notably, numerical algorithms find other possible solutions, as discussed in more detail in the results section (see \ref{sec:results}).

We use binary sequences as input to the auto-associator. To precisely define what we refer to as \textit{items} or \textit{relations}, the concept of a {\bf symmetry group} is essential; for symmetry related to geometry \cite{neumann1994groups}. The symmetry group of a set is a group of operations that leave this set invariant. Symmetry groups play an important role in many aspects, including linking algebra and geometry in mathematics, characterizing structures in crystallography, and their immediate relation with conservation laws in physics. 

First, we specify the group $\cal G$ of all considered admissible operations. Being concerned with binary sequences, a possible choice is to consider the group of all permutations on these sequences, i.e., the symmetric group $S_N$, where $N$ is the length of the patterns. Note that other choices are possible, displaying other coupling geometries \cite {reimann1998design}. The {\bf action of a group} element $s$ on an element $x = (x_i)$ in $X$ may be defined by permuting indices, i.e. $(s\cdot x)_i = x_{s(i)}$. The symmetry group of $X$ is the set of all permutations that leave the set $X$ invariant, i.e. Under the action of a permutation $s$ an element $x$ may be unchanged $s \cdot x = x$, or $x$ might be mapped to another element in $X$, i.e., $s\cdot x \in X$.
Under the action of its symmetry group, the set $X$ is partitioned into orbits. The {\bf orbit} of a point $x$ under the action of all elements $s$ of the symmetry group is the set of points that arise as actions of the elements of the symmetry group. We write $[x]=[x]_\Sigma$ for this orbit. Thus, two points $x,x'$ are on the same orbit $[x]$ if there is an element $s \in \Sigma_X$ such that $s \cdot x = x'$. Each orbit represents the set of symmetry-related items in $X$.

An auto-associator has to respect the symmetry structure of the training set in the sense that applying this transformation to the training set does not break or violate the underlying symmetries represented by that group. Formally: 
\begin{equation}\label{eq:compatible}
    \varphi_W(s \cdot x) = s \cdot \varphi_W(x)
\end{equation} 
for all elements $s$ in the symmetry group $\Sigma_X$ of $X$, where the symmetry group acts on the set of network matrices according to $(s \cdot W)_{ij} = W_{s(i)s(j)}$. In this case, the mapping $\varphi_W$ and the symmetries $\Sigma_X$ in the training set are compatible, and $\varphi_W$ is called $\Sigma_X$-compatible.

{\it $\varphi_W$ is $\Sigma_X$-\textit{compatible} if and only if $W$ is invariant under the action of the symmetry group}, i.e.
\begin{equation}
 W = s \cdot W    \qquad \mbox{for all $s \in \Sigma_X$}
\end{equation}

Therefore, the condition that $\varphi_W$ is compatible with the symmetry structure of the training set is a condition on the network structure. 
The network structure $W$ must be a representation of the symmetry group $\Sigma_X$ of the training set. In this case, the network represents the relations in the training set. When factoring out the symmetry, what remains is the set of orbits. Thus, by using the symmetry-adapted network, the problem boils down to solving the fixed point equations on each orbit, i.e., $[x] = \varphi_W([x])$ for all orbits $[x]$. 
The following characterizes an auto-associator acting on a training set $X$: 

{\it $\varphi_W$ is an \textit{auto-associator} on $X$ if it is $\Sigma_X$-{compatible} and if the fixed point equations are fulfilled on each {orbit} under $\Sigma_X$, i.e.}

Under the action of its symmetry group $\Sigma$, the training set $X$ is partitioned into orbits $[x]$, all points in one orbit being equivalent. It follows that if one point in the orbit, e.g., $x$ itself, is stable, then all other points belonging to $[x]$ are also stable \cite{reimann1998design}.


\section{Analytical and numerical results}
In the following, we apply the theory to two training sets consisting of two training items each. We calculate the respective auto-associative mappings explicitly. This allows us to compare these analytical solutions with the numerical solutions found, see below. 
\begin{equation}\label{eq:training_items_ex1}
    X = \left\{ 
    \begin{pmatrix} 1\\0\\1\end{pmatrix}, 
    \begin{pmatrix} 1\\1\\0\end{pmatrix}
    \right\} \hskip 1cm
    X' = \left\{ 
    \begin{pmatrix} 0\\1\\0\end{pmatrix}, 
    \begin{pmatrix} 0\\0\\1\end{pmatrix}
    \right\}
\end{equation}

Notably, both training sets have the same symmetry group, $\Sigma_X = \{e,(32)\}$, where $e$ is the identity and $(32)$ permutes the second and the third element. The action of a group element $s$ on an element $x = (x_i)$ in $X$ is defined as $(s\cdot x)_i = x_{s(i)}$. For example $(e\cdot x)_2 = 2$, while $((32)\cdot x)_2 = 3$. Since both training sets have the same symmetry group, the respective symmetry-compatible networks are identical. According to eq. \ref{eq:compatible}, the $\Sigma_X$-compatible network is
\begin{equation}\label{eq:W}
    W = \begin{pmatrix}
        a&b&b\\
        c&d&e\\
        c&e&d
    \end{pmatrix}
\end{equation}
The auto-associative network has to respect this structure. We now explicitly calculate the solutions of the auto-associative task $\varphi(W\:x)=x$ for test items $x \in X$ and $\varphi(W'\:x)=x$ for test items $x \in X'$.

\subsection{Linear auto-associator}
First, consider the linear activation function $\varphi(x)=x$. Since orbits are different, so are the fixed point equations. Therefore, the resulting auto-associative networks are different:
\begin{equation}\label{eq:WAA}
    W_{a,b} = \begin{pmatrix}
        1-a&a&a\\
        -b&1+b&b\\
        -b&b&1+b
    \end{pmatrix} \hskip 1 cm
    W_{a,b}' = \begin{pmatrix}
        a&0&0\\
        b&1&0\\
        b&0&1
    \end{pmatrix}
\end{equation}
where $a,b$ are arbitrary parameters. 
Note that both networks have the symmetry-derived structure given in eq. \ref{eq:W}. Respective eigenvalue spectra are $\sigma_W = \{1-a-2b,1,1\}$ and $\sigma_{W'} = \{a,1,1\}$. Therefore, stability requirements on orbits impose an additional constraint on the parameters. 

\subsubsection{Numerical solutions of network structure depend on network hyperparameters}
Next, we aim to compare our analytical results to numerical solutions obtained with commonly used ANN optimization algorithms. Specifically, we train a linear auto-associator to represent the two training items shown in \ref{eq:training_items_ex1} minimizing the mean squared error between the input $x$ and the generated output ${\hat x}=\varphi_W(x)$:
\begin{equation}\label{eq:rnn_cost}
loss = \frac{1}{N}\Sigma_{i=1}^N({\hat x}_i - x_i)^2
\end{equation}
We use two well-known algorithms such as (Stochastic Gradient Descent (SDG \cite{sutskever2013importance}) and ADAM optimizer \cite{kingma2014adam}). Additionally, we initialized the coupling matrix $W$ as either 
homogeneously $0$, homogeneously $1$, or random using either a uniform ($\mathcal{U}(0,1)$) or a Gaussian distribution ($\mathcal{N}(\mu=0, \sigma=0.05)$). The trained networks are indexed according to the algorithm used and the initialization method of the coupling matrix $W$. Under all conditions, the networks learned to represent the training items exactly (training cost = 0; see eq. \ref{eq:rnn_cost} and table \ref{tab:loss}).

\begin{eqnarray*}
    && W_0^{SDG} = \begin{pmatrix}
        0.6667 & 0.333 & 0.333\\
        0.3333 & 0.6667 & -0.3333\\
        0.3333 & -0.3333 & 0.6667
    \end{pmatrix} \quad 
    W_0^{ADAM} = \begin{pmatrix}
        0.5496 & 0.45004 & 0.4504\\
        0.4995 & 0.5005 & -0.4996\\
        0.4996 & -0.4996 & 0.5004
    \end{pmatrix} \\
    && W_1^{SDG} = \begin{pmatrix}
        0.3333 & 0.6667 & 0.6667\\
        0.0000 & 1.0000 & 0.0000\\
        0.0000 & -0.0000 & 1.0000
    \end{pmatrix} \qquad 
    W_1^{ADAM} = \begin{pmatrix}
        0.4504 & 0.5496 & 0.5496\\
        0.0522 & 0.9478 & -0.0522\\
        0.0489 & -0.0489 & 0.9511
    \end{pmatrix}\\
    && W^{SGD}_{rnd} = \begin{pmatrix}
    0.5068  &  0.4932  &  0.4932\\
    0.1697 &   0.8303  & -0.1697\\
    0.2098  & -0.2098  &  0.7902
    \end{pmatrix}\hskip 0.45cm 
    \mathbf{E} W_{rnd}^{SGD} = \begin{pmatrix}
        0.6668 & 0.3328 & 0.3329\\
        0.3332 & 0.6672 & -0.3328\\
        0.3332 & - 0.3328 & 0.6671
    \end{pmatrix}
\end{eqnarray*}

The obtained solutions are numerically different but exhibit the common structure displayed in eq. \ref{eq:WAA}. All numerical solutions are particular members of the 2-parameter family of solutions $W_{a,b}$. Particularly $W_0^{SDG}$ has parameters $b=\frac{1}{3}$ and $c=b$, $W_1^{SDG}$ has $b=\frac{2}{3}$ and $c=0$, $W_0^{adam}$ has $b=0.4504$ and $c=0.4995$, while $W_1^{adam}$ has $b=0.5496$ and $c=0.0489$. 

A third setting to be considered is that the network matrix is randomly initialized, i.e., weights are randomly drawn from a Gaussian distribution ${\cal N}(0,1)$. Trained networks resulting from random initializations differ. Examples are 
\begin{equation*}
W^{SGD}_{rnd} = \begin{pmatrix}
0.6367  &  0.3633  &  0.3633\\
    0.3659  &  0.6341  & -0.3659\\
    0.3194 &  -0.3194 &   0.6806
\end{pmatrix}, \;
W^{SGD}_{rnd} = \begin{pmatrix}
0.5068  &  0.4932  &  0.4932\\
    0.1697 &   0.8303  & -0.1697\\
    0.2098  & -0.2098  &  0.7902
\end{pmatrix} \hdots
\end{equation*}
The resulting realizations of auto-associative matrices are different but exhibit a common structure that fulfills the fixed point equation. 
$W = \begin{pmatrix}
        1-a &a&a\\
        -b&1+b&b\\
        -c & c & 1+c
    \end{pmatrix} $.
However, note that averaging over realizations recovers the symmetry solution eq. \ref{eq:WAA}. Particularly compare 
\begin{equation*}
    \mathbf{E} W_{rnd}^{SGD} = \begin{pmatrix}
        0.6668 & 0.3328 & 0.3329\\
        0.3332 & 0.6672 & -0.3328\\
        0.3332 & - 0.3328 & 0.6671
    \end{pmatrix} \quad 
    \mathbf{E} W_{0}^{SGD} = \begin{pmatrix}
        0.6667 & 0.3333 & 0.3333\\
        0.3333 & 0.6667 & -0.3333\\
        0.3333 & - 0.3333 & 0.66667
    \end{pmatrix}
\end{equation*}

\subsubsection{Linear auto-associator generalizes to items outside the training set}
In linear auto-associators, the symmetry group of the training items is also reflected in the network function.
Specifically, the linear auto-associator not only reproduces the elements it has been trained but also generalizes to represent two binary patterns additional to the training set with a $loss=0$ (see table \ref{tab:loss}). Notably, the property of $W$ to be an auto-associator on a set $X$ generalizes to the set of all elements $y$ such that the extended set $X' = X \cup \{y\}$ also has the symmetry group $\Sigma_X$. This is $W$, which acts as an auto-associator on those elements $y$, which are 'compatible' with the symmetry group of its training set. 

Note that the linear auto-associator $W$, which has been designed or trained on some training set $X$, i.e., $x = Wx$ for all $x\in X$, also acts as an auto-associator on all patterns which are linear combinations of the elements in the training set: Let $y = \sum_{x \in X} a_x \: x$ be a linear combination of training examples $x$ and $\alpha_x$ its coefficient. Then $Wy = W \sum_{x \in X} a_x \: x = \sum_{x \in X} a_x \: W x = y$, i.e. $y$ is an invariant under $W$. 
Furthermore, the entire plane $V$ spanned by the elements in the training set $X$ is invariant under $W$. If this attractor is stable, arbitrary inputs will be mapped in that plane after an infinite iteration, i.e., $W^nx_0 \to x_\infty \in V$ plane attractor. 
%
\begin{table}[h] 
\centering
\begin{tabular}{c|c||c|c}
\textbf{$x$} & $\mathbf{id}$ & {$\mathbf{tanh}$} & $\mathbf{sigmoid}$ \\ \hline
0/1/0      & 0.0000  & 0.0001 & 0.0003  \\
0/0/1      & 0.0000  & 0.0001 & 0.0003  \\ \hline
0/0/0      & 0.0000  & 0.0000 & 0.2500  \\
0/1/1      & 0.0000  & 0.0002 & 0.1663  \\
1/1/0      & 0.4251  & 0.3008 & 0.3099 \\
1/0/1      & 0.4251  & 0.2459 & 0.3100  \\
1/1/1      & 0.4251  & 0.1751 & 0.4043  \\
1/0/0      & 0.4251  & 0.3716 & 0.3330  \\ \hline
\end{tabular}
\vspace{6pt}
\caption{\textit{Generalisation} The table shows the cost \ref{eq:rnn_cost} on training and test sets for linear and non-linear activation functions. Particularly, $\phi_W(x) = Wx$, $\varphi_W(x)=\tanh(Wx)$, and $\varphi_W(x) = \frac{1}{1+exp(-Wx)}$. The first two items are the training items and are correctly reproduced by all networks considered. Networks differ in their generalization ability: The linear network reproduces the foremost 'unseen' items $(0,0,0)$ and $(0,1,1)$. At the same time, they fail to be reproduced by the non-linear network with a sigmoid activation function.}\label{tab:loss}
\end{table}

To better understand how the networks can achieve this generalization, we studied the spectra of the coupling matrix $W$.
Recall that the spectra of the analytically derived network solutions $W$ are $\sigma_W = \{1-a-2b,1,1\}$ and $\sigma_{W'} = \{a,1,1\}$ so that the fixed point $\begin{pmatrix}1\\0\\1 \end{pmatrix}$ together with all points on its orbit is asymptotically stable if $|a-2b| < 1$. In this case, the plane spanned by $X$ is an asymptotically stable attractor. Analogously, the point $\begin{pmatrix}0\\1\\0 \end{pmatrix}$ is asymptotically stable if $|a|<1$. In this case, $X'$ spans a stable plane attractor. 

In line with our analytical results, all trained coupling matrices $W$ had two eigenvalues $\lambda_1=\lambda_2=1$ and one eigenvalue $\lambda_3 << 1$. The corresponding first and second eigenvectors span a plane that goes through the origin (0/0/0) and contains the two elements of the training set. The third eigenvector was roughly orthogonal to this plane. Hence, the networks set up a plane attractor so that any (also non-binary) element $x$ located on this plane will be exactly reproduced by the network. In contrast, all elements outside this plane are pulled towards the plane ($\lambda_3 << 1$) and are therefore not mapped onto themselves by the network (loss > 0). Furthermore, every linear combination of the training elements is located on this plane, demonstrating how linear auto-associators can reproduce not only the training elements but any linear combination of them. 

\begin{figure}[H]
\hspace{-41pt}
\includegraphics[width=1.2\linewidth]{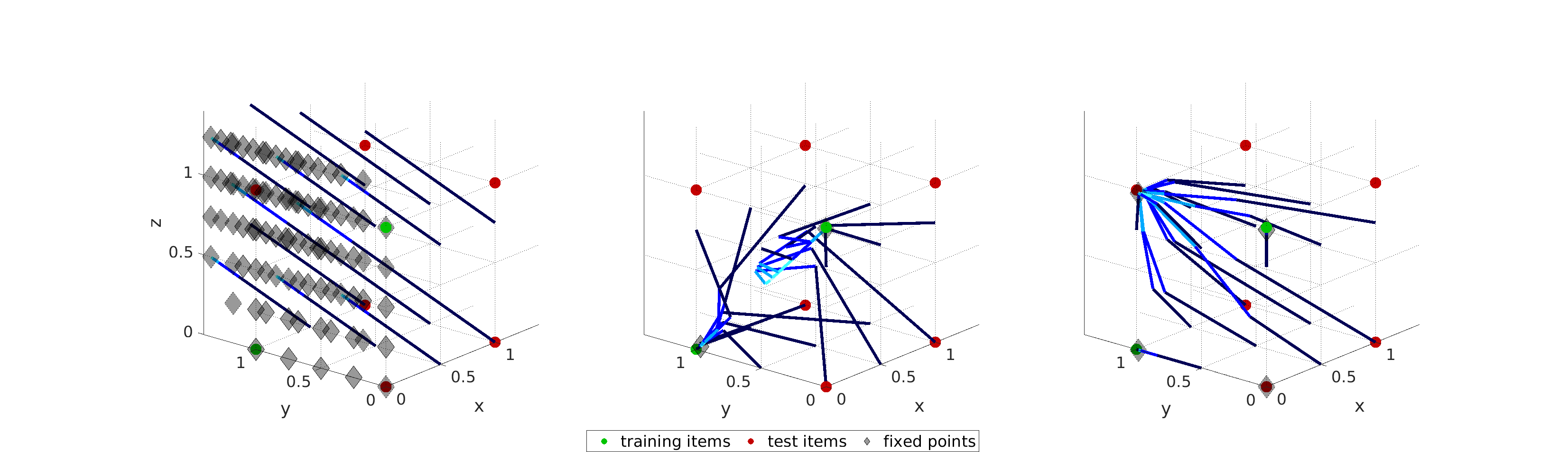}
\caption{\textit{Network dynamics across activation functions}
Evolution of 18 input elements $x$ for repeatedly applying W for networks with $linear$ (left), $sigmoid$ (middle), and $tanh$ (right) activation function $\varphi$ (dark blue line shows flow field at n=0, light blue line at n=6; fixed points defined as $x$ at $n=25$).
These extended flow fields illustrate how the linear network sets up a plane attractor and, therefore, is able to represent any input element located on this plane. In contrast, the non-linear networks implement fixed points and show a lower generalization ability.} \label{fig:network_dynamics}
\end{figure}

These plane attractor dynamics are illustrated in Fig.~\ref{fig:network_dynamics}. To visualize the corresponding flow field, we place 125 (non-binary) elements equally spaced on a 3-dimensional mesh, spanning the network's input (and output) space. Then, we apply the network function n-times to each element and observe the trajectory of each element to study the network dynamics (only 18 trajectories are shown for visibility). The blue-colored lines in Fig.~\ref{fig:network_dynamics} describe the evolution of different (also non-binary) input elements $x$ ($W^nx$; dark blue $n=0$, light blue $n=6$). As expected, based on the studies of the network spectra, all trajectories converge onto a plane spanned by the two training items. 

Overall, a linear auto-associator seems to learn the relation of the training elements. 
While the exact entries of the trained coupling matrices $W$ depend on the training algorithm and the initialization, the connectivity reflects the symmetry group of the training set. Similarly, the network dynamics implement a plane attractor, which enables the network to solve the auto-association task for unseen elements of the same symmetry group, too.


\subsection{Non-linear auto-associator}
A non-linear auto-associator has fundamentally different properties. Its non-linearity is due to the non-linearity of some activation function $\varphi$. Here we consider two activation functions: $\varphi(x) = \tanh(x-c)$ and $\varphi(x)= \frac{1}{1+e^{-c\:x}}$ the sigmoid. 

\subsubsection{Non-linear auto-associator's structure represents training items' symmetry group}
As in the linear case, the network $W$ again has the structure inherited from the symmetry group of the training set. Intuitively, the reason is that the non-linear activation function is still monotonous, thereby conserving the relations. The network of the non-linear auto-associator equals that of the linear one, except that it has an additional term accounting for the non-linearity. 
\begin{equation}
    W_{abc} = W_{ab} + W_c,
\end{equation}
where $W_{ab}$ is the network structure of the linear auto-associator, see eq. \ref{eq:WAA}. For the 'tanh' activation function, the additional term respecting the non-linearity yields
\[
W_{c} = \begin{pmatrix}
\tanh(1+c)-1&0&0\\
\tanh(c)&\tanh(c)-\tanh(1+c)+1&0\\
\tanh(c)&0&\tanh(c)-\tanh(1+c)+1
\end{pmatrix}
\]
Similarly, for the 'sigmoid' activation function, the term is 
\[
W_{c} = \begin{pmatrix}
-\frac{e^{-c}}{1+e^{-c}}&0&0\\
\frac{1}{2}&\tanh(c)-\tanh(1+c)+1&0\\
\frac{1}{2}&0&\frac{1}{2}\frac{1-e^{-c}}{1+e^{-c}}-1
\end{pmatrix}
\]
Note that $W_c$ has the same structure as $W_{ab}$.

\subsubsection{Non-linear auto-associator does not generalize to the symmetry group of training items}
Unlike the linear network, some non-linear auto-associators can only reproduce the elements of the training set. 
Specifically, non-linear auto-associators using a sigmoid activation function can only represent the training items. In contrast, auto-associators using the tanh activation function generalize to identical input elements as linear networks. However, the resulting loss values are slightly larger than in the linear case (see table \ref{tab:loss}).

To understand how the networks achieve these different results, we again study the network dynamics. However, compared to the linear case, the eigenvalue decomposition of the coupling matrix $W$ is no longer easily interpretable as it does not reflect the effect of the non-linear activation function $\varphi$. Therefore, we study the network dynamics by examining the network's flow field (see Fig.~\ref{fig:network_dynamics}). Since the network dynamics operate in a 3-dimensional state space, we can visualize the full dynamics. Similar to the linear case, we place 125 (non-binary) elements equally spaced on a 3-dimensional mesh and apply the network function n-times to each element. This allows us to observe how the network dynamics affect different input elements $x$.

This analysis reveals that the two non-linear networks implement different solutions to solve the auto-associator task.
Auto-associators with a sigmoid activation function realize an associative memory on the training set, i.e., the elements of the training set are attractive fixed points. The flow field shows that all input elements fall first onto the line span between the two training items and eventually converge to either of the two training items. 
In contrast, the tanh network sets up four fixed points - one for each input element it can represent. However, only one fixed point is attractive, and all probed input elements converge to the fixed point at $(0,1,1)$. The tanh network's linearized dynamics are remarkably similar to those of the linear network. It consists of two eigenvectors in the plane span by the two training items. However, unlike in the linear case, the corresponding eigenvalues are unstable ($\lambda_1=2.43, \lambda_2=  2.43$). Hence, the network has a strong drive to push any input elements away from the origin on this plane. This is beneficial for the network as it has to achieve values of $1$ to represent the tested binary input elements, which - given the tanh non-linearity - can only be achieved with $x$ >> 1.
The difference between networks with these two activation functions can be partially explained by looking at the ranges in which they are strongly non-linear for binary input elements: for the sigmoid, these are the values around $0$ and $1$, while the tanh is only non-linear close to $1$ and remains almost linear in the neighborhood of $0$. 

Overall, these results show that non-linear networks tend to learn items, not relations. However, suppose the non-linearity is such that that network operates in a nearly linear regime (e.g., tanh around $0$). In that case, networks can at least partially generalize to elements outside the training items similar to a linear network. 


\section{Discussion}
Modern operative ANNs typically exhibit massive input and network parameter dimensions, while interactions within layer architecture are non-linear. This provides many degrees of freedom, which allows them to cope with extended data and complicated tasks. On the other hand, the same complexity makes it challenging to understand what is going on when networks learn to fulfill their respective tasks. Being interested in understanding what is learned, we, therefore, reduced the setting to the probable most simple stage:  We chose a simple architecture, a two-layered network, one layer for input, the other one for output, which moreover is small. Furthermore, we chose a simple task: Identically reproduce a set of training items. Neurons are classical perceptrons coupled all-to-all. Variations considered are due to the choice of different activation functions, each of which is standard, linear, or non-linear (e.g., sigmoid or tanh). This framework is simple enough to allow for analytical solutions to the auto-associator task. This theory applies to arbitrary network size and also generalizes to more complicated network architectures. We used simple examples for which we obtained concrete analytical results. Knowing the analytical truth about these examples, we could judge numerical results and understand why they look as they do. Our main statements are highlighted in the following.

\paragraph{Numerical solutions are correct but different.}
We used standard algorithms such as the basic delta rule, SGD, and ADAM to let the networks learn from different initial conditions. While learning, the network changes structure until it correctly solves the task. Algorithms used produced correct solutions. Interestingly, the trivial solution, i.e., the connectivity matrix being the unity matrix, is {\it not} found numerically. Moreover, solutions depend on the particular algorithm used and the initial network structure. While this is expected in high dimensions, our setting is just 3-dimensional. 
It turns out that numerical solutions are particular members of an infinite family of solutions.

\paragraph{The structure of the auto-associator network represents the structure of the training set.} 
For concreteness, we restricted ourselves to binary sequences and the action of permutations on them. The structure of the training set is described by its symmetry group, i.e., the set of all permutations acting on binary sequences, which leave this set invariant. We showed that the structure of the auto-associator network and the structure of the training set are closely related: The connectivity matrix of the auto-associator network represents the symmetry group of the training set. All numerical solutions share the same structure. Factoring out the symmetry from the training set reduces the problem of finding fixed points to the representatives of the orbits. 
By conducting this idea, we were able to calculate the auto-associative mapping and solve the task. This analytical solution allowed us to compare numerical solutions.


\paragraph{Linear auto-associators generalize, while non-linear auto-associators learn items.}
Recall that in our setting, {\it items} are the binary sequences in the training set, and {\it relations} are the permutations that leave it invariant. Notably, we asked what the network has learned: items or relations between these items. Our derivations show that an auto-associator structure must represent the structure of the training set (i.e., the relations between training items) to generalize. While linear and non-linear network weights $W$ represent the symmetry structure, only the linear auto-associators could generalize, and it did so by implementing a plane attractor. In contrast, non-linear auto-associators resemble more associative memory and have learned to represent the training items using stable fixed points. 




\paragraph{Regularization improves generalization if the activation function admits a linear regime.}
L1, L2, or firing rate regularization are often employed along with other methods to achieve a more robust and generalizable model. They add a penalty term proportional to the (squared) magnitude of the model's coefficients (firing rate) to the cost function. Thereby, they penalize large values in the parameter vector, encouraging the model to use smaller weights and activation values. In the case of the tanh-activation function, this becomes relevant because it lets the network operate in two different regimes: for values around 0, the regime is almost linear, while for large values, it is non-linear. Applying any of these regularizations thus potentially pushes the network closer to or even into this almost linear regime.
Consequently, these regularizations for the system with a tanh-like activation function generate some degree of generalization. The network with the sigmoid activation will not benefit directly from this regularization. Instead, it would require a regularization that enforces network activities to stay around 0.5 (linear regime of sigmoid function).

\paragraph{ARC as an example}. Learning and representing relations in a training set are of particular importance in the so-called Abstraction and Reasoning Challenge (ARC): Each task contains 3-5 pairs of pictures displaying training inputs and outputs, which are related by the same but unknown construction rule, also called a concept. A test input is then provided, and the agent is asked to predict the 'corresponding' output, i.e., a pattern constructed according to the same rule(s) as the set of training pairs. The challenge, therefore, is bi-fold: First, infer a set of those rules, actually a maximal set, transforming each input to its training output; Second, apply these rule(s) to the test item to construct the resulting output. 

\begin{figure}[h]
    \includegraphics[scale=0.4]{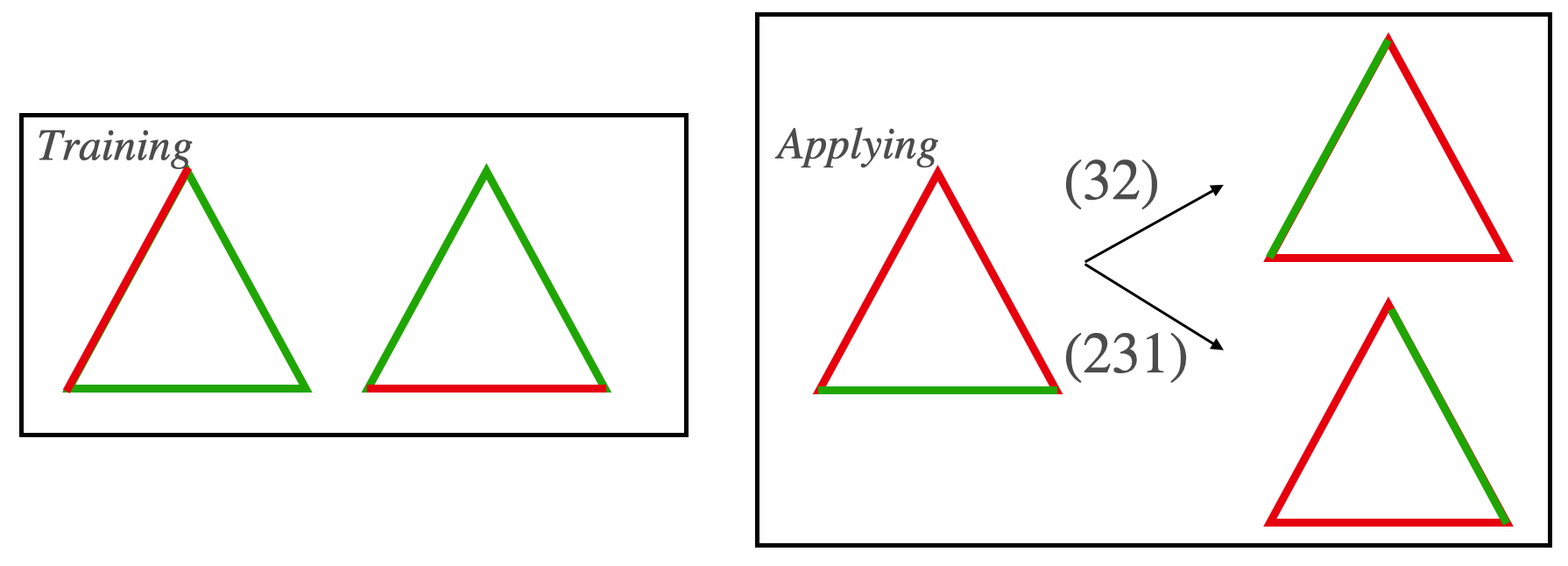}
    \caption{\textit{Symmetry example} $(32)$ is a symmetry operation on the training set, while the rotation $(231)$ is not.}\label{fig:SolvedARC2}
\end{figure}

The figure below shows two triangles with colored edges, visually displaying two binary sequences of length $3$. Each triangle represents a training item in $X'$: $0$ codes for a green edge while $1$ codes for a red one. The test triangle corresponds to the pattern $\begin{pmatrix}1\\1\\0 \end{pmatrix}$. The only non-trivial invariant rule inferred from the training set is to interchange the second and the third edge, corresponding to the permutation $(32)$. Applying this rule to the test pattern/triangle results in $\begin{pmatrix}1\\0\\1 \end{pmatrix}$, see Figure \ref{fig:SolvedARC2}.  

In our approach, the group of operations is given (e.g., all permutations on sequences of $N$ elements). The network's generalization then is due to its subgroup of symmetry operations. What has to be learned is the set of operations large enough to contain a sufficiently big symmetry group of the training set. This shifts the requirement 'Learn items!' towards the requirement 'Generate a sufficiently rich repertoire of operations!'

\section{Acknowledgements}
We thank V. Mante for supporting us in this project. This work was funded by the Simons Foundation (SCGB 328189 and 543013, V.M.) and the UZH Postdoc Grant (FK-22-116, R.K.).

\bibliography{bib_symGroups.bib}

\end{document}